\documentclass[runningheads]{llncs}
\usepackage[T1]{fontenc}
\usepackage{graphicx}
\usepackage{subfigure}

\usepackage{hyperref}

\usepackage[utf8]{inputenc}
\usepackage{amsmath}
\usepackage{amssymb}
\graphicspath{{figs/}}
\usepackage{duckuments}
\usepackage{cleveref}
\usepackage{comment}
\usepackage{array}
\usepackage{tikz}
\usetikzlibrary{calc}
\renewcommand{\Re}{\mathbb{R}}
\newcommand{\Ze}{\mathbb{Z}}
\renewcommand{\vec}[1]{\mathbf{#1}}
\newcommand{\mat}[1]{\mathbf{#1}}
\newcommand{\vecop}[1]{\operatorname{vec}\!\left({#1}\right)}

\definecolor{jon}{rgb}{0.79,0.25,0.33}
\definecolor{sune}{rgb}{0.94,0.88,0.18}
\definecolor{peidi}{rgb}{0.0,0.0,0.5}
\definecolor{lu}{rgb}{0.73,0.72,0.42}
\definecolor{tchoa}{rgb}{0.31,0.09,0.61}

\begin{document}
\title{Locally orderless networks}
%
%
\author{Jon~Sporring\inst{1,2}\and
Peidi~Xu\inst{1} \and
Jiahao~Lu\inst{1} \and
François~Lauze\inst{1,2} \and
Sune~Darkner\inst{1}}
\authorrunning{Sporring et al.}
%
\institute{Department
of Computer Science, University of Copenhagen, Denmark
\email{\{sporring,peidi,lu,darkner\}@di.ku.dk} \and
Center for Quantification of Imaging Data from MAX IV (QIM), Denmark}
\maketitle              
\begin{abstract}
We present Locally Orderless Networks (LON) and its theoretic foundation which links it to Convolutional Neural Networks (CNN), to Scale-space histograms, and measurement theory. The key elements are a regular sampling of the bias and the derivative of the activation function. We compare LON, CNN, and Scale-space histograms on prototypical single-layer networks. We show how LON and CNN can emulate each other, how LON expands the set of functionals computable to non-linear functions such as squaring. We demonstrate simple networks which illustrate the improved performance of LON over CNN on simple tasks for estimating the gradient magnitude squared, for regressing shape area and perimeter lengths, and for explainability of individual pixels' influence on the result. 
\end{abstract}

\section{Introduction} 
We introduce the locally orderless network (LON) which locally transforms the input signal into a set of local histograms. We bring a novel perspective on activation functions and scale-space which apply learnable operators to form linear combinations of measurable sets in order to extract relevant information. This is a generalization of the Locally Orderless Image (LOI) paradigm \cite{koenderink99} which introduces a scale-space on the image, integration, and intensity space making histograms smooth in all dimensions.  The LON layer introduces the classic scale-space into neural networks (NN) with a strong theoretical foundation from scale space and measure theory and proposes a novel perspective on many of the classic concepts of invariance and density estimates in neural networks. Our purpose is to understand the fundamental capacities of the LON and compare it to Convolutional Neural Networks (CNN)~\cite{fukushima80}. Thus we describe the LON from a theoretical perspective and compare it with a similar CNN on simple binary examples to illustrate some of their properties.

\section{Previous Work}
The LON produces densities which are estimated through the locally orderless framework originating from LOI, rooted in halftones ideas and the fact that the contents of an image can still be identified if a local spatial permutation of the intensities is performed. 
The locally orderless theory exposes the 3 inherent scales of density estimation: the resolution of the signal, the size of the local neighbourhood, and the bin width in terms of Gaussian convolutions. The theory has mainly been applied to image similarity and image registration where the different estimators have been compared \cite{hermosillo02} and a generalized formulation of image similarity in image registration have been introduced \cite{darkner.sporring13}. The formulation was then extended to a scale-space model for DWI geometry \cite{jensen15,jensen21} and higher-order information \cite{darkner.vidarte.lauze21}. The density estimation is key to estimating a wide range of similarity measures within the LOI framework and the similarity measures can be classified into linear and non-linear similarities like the linear generalized P-Norm and the non-linear measures such as mutual information, and cross-correlation, etc. 

To the best of our knowledge, there is little prior work that is closely related to the LON. Histograms have been used in networks previously, where \cite{sedighi2017histogram} introduced the histogram layer for more compact representation and realized that this is induced by the activation function. In \cite{hussain2019imhistnet} a global version was introduced, and \cite{pmlr-v139-list21a} showed how such a representation can be used to estimate quantiles and subsequently distance. We connect all these contributions in the LON framework, where it becomes obvious that the difference is merely a choice of measure on the sets and the scale of integration. Density estimation is no stranger to neural networks and machine learning in general and many models are formulated in terms of probabilities. Work within deep learning network components exists, but most of that relates to density for loss functions such as cross-entropy. The only function that is remotely related to histograms and density operation is the pooling operation that can be interpreted as a histogram operation. The learning of pooling operations was suggested by\cite{SUN201796} where the pooling operation was learned during training. As an alternative one could consider Bayesian Neural Networks \cite{bishop97} that work with parameter uncertainty or distributions. In contrast, LON works with distributions of data.  There are however some works before the breakthrough of neural networks. In \cite{ginneken00,ginneken03} where local histograms were used for texture classification with success. Furthermore, many of the classical image descriptors such as histogram of Gaussians (HoG), Daisy \cite{tola.lepetit.fua10}, and SIFT \cite{Lowe1999} are all based on histograms of features in some way.  
  
\section{Locally orderless histograms as convolutional networks}
Consider an image and two kernels $I, K, W:\Re^d\rightarrow\Gamma$, with support in $\Omega\subseteq\Re^d$, and $\Gamma\subseteq\Re$, a function $f:\Re\rightarrow\Re$, and scalars $b\in\Re$, and the function $h:\Omega\times\Re\rightarrow\Re$,
\begin{align}
h(\vec x, b) 
&= \int_{\Re^d} f\!\!\left(b-\int_{\Re^d} I(\vec \beta-\vec \alpha)K(\vec \alpha)\, d\alpha\right) W(\vec x -\vec \beta)\,d\beta, 
\label{eq:localHistogram}
\end{align}
which we denote as $h(\vec x, b) = \left(W * f\!\left(b-\left(I*K\right)\right)\right)\!\left(\vec x\right)$. When $W$ is a smoothing kernel, and $f$ is a bell shaped function whose integral is a sigmoid function, then $h(\vec x, b)$ is a the local histogram value of $I*K$ in the neighbourhood of position $\vec x$ and intensity $b$. In \cite{koenderink99}, the functions $W$ and $K$ are Gaussian, $f$ is an unnormalized Gaussian with $f(0)=1$, and all 3 functions had independent width parameters. Images described by $h$ are called locally orderless images (LOI).

In the rest of this article, we use a standard discretization of space such that $\Omega = \{\vec x_k\}\subset\Ze^d$. We introduce a locally orderless network layer for a discrete inputs $I_k = I(\vec x_k)$ as linear combination of $M$ local histograms $h_j$ with individual kernels $K_j$ and $W_j$ and $2NM$ kernel dependent intensities $b_{ij}$ and bell shaped functions parametrized by $\sigma_{ij}$, as $\text{LON}:\Re^{|\Theta|}\times\Gamma^{|\Omega|}\rightarrow\Re^O$,
\begin{align}
h_{ijk} &= \left(W_j * f_{\sigma_{ij}}\!\left(b_{ij}-\!\left(I*K_j\right)\right)\right)\!\left(\vec x_k\right),\\
\text{LON}\left(\{I_k\}\right) &=\mat{A}\vecop{\{h_{ijk}\}}
\label{eq:lon}
\end{align}
where $\Theta = \{K_j,b_{ij},\sigma_{ij},W_j,\mat{A}: i = 1\ldots N, j=1\ldots M\}$ is the set of parameters, $\mat A\in\Re^{MN|\Omega|\times O}$ is a matrix and $O$ is the number of output connections under suitable boundary conditions for the convolution operator, and $|\cdot |$ is the cardinality operator.

In the following, we will give an interpretation of LON by comparing it with a similar convolutional network, and we will show how LON can calculate a large class of non-linear functionals. 

\subsection{Boundaries versus areas}
A convolutional network similar to \eqref{eq:lon} is found by replacing the bell shaped functions $f_{ij}$ with a single sigmoid function $g_{ij}(v) = \int_{-\infty}^v f_{ij}(w)\,dw$,
\begin{align}
\text{CNN}\!\left(\left\{I_k\right\}\right) &= \mat{A}\vecop{\!\left\{\left(W_j * g_{ij}\!\left(b_{ij}-\!\left(I*K_j\right)\right)\right)\!\left(\vec x_k\right)\right\}}.
\label{eq:cnn}
\end{align}
Consider a family of activity functions in LON which converges to Kronecker's delta function $f_{\sigma_{ij}}\rightarrow\delta$, as $\sigma_{ij}\rightarrow 0$, then correspondingly, the activity functions in CNN will converge to the Heaviside function, $g_{ij}\rightarrow H$. As a consequence, 
\begin{align}
f_{ij}\!\left(b_{ij}-\!\left(I*K_j\right)\!\left(\vec x\right)\right) &\rightarrow 
\begin{cases}
    1, \text{ if } b_{ij}-\!\left(I*K_j\right)\left(\vec x\right) = 0\\
    0, \text{ otherwise}
\end{cases}\\
g_{ij}\!\left(b_{ij}-\!\left(I*K_j\right)\!\left(\vec x\right)\right) &\rightarrow 
\begin{cases}
    1, \text{ if } b_{ij}-\!\left(I*K_j\right)\left(\vec x\right)\geq 0\\
    0, \text{ otherwise}
\end{cases}.
\end{align}
Thus LON focuses on the isophotes of $b_{ij}-\!\left(I*K_j\right)\!\left(\vec x\right)$, while CNN performs a threshold of the same term. In the limit of $f\rightarrow\delta$, the isophote is ill-posed, however, as discussed in \cite{koenderink99}, when $f$ has a finite width, then the result of applying it as in LON defines well-posed, soft isophotes.

A consequence of the above is that we can design a very economical LON for estimating circumferences and a very economical CNN for calculating areas. Given a connected, compact region $S\subset\Omega$ and an image $I = 1(S)+\varepsilon$, where $1$ is the indicator function and $\varepsilon$ is i.i.d.\ noise. Then
\begin{align}
\text{Circumference}(S) \sim \sum_k \left(\delta * f\!\left(0.5-\!\left(I*K\right)\right)\right)\!\left(\vec x_k\right),\\
\text{Area}(S) \sim \sum_k \left(\delta * g\!\left(0.5-\!\left(I*K\right)\right)\right)\!\left(\vec x_k\right),
\end{align}
where $K$ is a smoothing kernel, and $\mat{A}$ are implied to be 1-vector with $|\Omega|$ elements. Both the circumference and area will vary proportionally to the degree of smoothing and the curvature of $S$ and the width of $f$ and $g$. 

\subsection{LON can implement nonlinear measures directly}
\label{sec:GradMagnitude}
The locally orderless network may be used to compute local image operators not easily computed with standard convolutional neural networks. 

For an transformation $\xi:\Gamma\rightarrow\Gamma$, $J(\vec x)=\xi(I(\vec x))$, and for a probability mass function $h_I$, by the Law of the unconscious statistician we have,
\begin{align}
    \mathbb{E}\left(J\right)=\sum_\Gamma \xi(i)h_I(i),
    \label{eq:unconscious}
\end{align}
A local version is obtained by convolution with a smoothing kernel $W$,
\begin{equation}
    \left(W*J\right)(\vec x) \simeq \sum_{i\in\Gamma}\xi(i)h(\vec x, b_i).
\end{equation}
Since the above is linear in $h$ then it can be written on the form \eqref{eq:lon}. Further, any linear combinations of transformations $\xi_m:\Gamma\rightarrow\Gamma$,
\begin{equation}
    \sum_m\left(W*J_m\right)(\vec x) \simeq \sum_m\sum_{i\in\Gamma}\xi_m(i)h(\vec x, b_i).
\end{equation}
is also linear in $h$ and thus can also be written on the form \eqref{eq:lon}.

As an example of using the the Law of the unconscious statistician, consider derivative kernels, $K_k(\vec x)$ where $\vec x = (x_1,x_2,\ldots,x_d)$, and such that $I_k=I*K_k$ is a smoothed estimate of the directional derivative in the $x_k$-direction. With $\xi(v) = v^2$, the LON can approximate the gradient magnitude squared as, 
\begin{align}
    \text{grad}^2(\vec x) \simeq \sum_{k=1}^d\left(W*I_k^2\right)(\vec x)
    &\simeq \sum_{k=1}^d\sum_{i\in\Gamma} i^2 h_k(\vec x, i),
\label{eq:grad3}
\end{align}
Note that for linear functions $\xi$ and Gaussian kernels $W$ and $K$ or its derivatives with standard deviation $\gamma$ and $\sigma$, then convolution semi-group property of Gaussian kernels implies that the two kernels can be replaced with a single Gaussian kernel of standard deviation $\sqrt{\gamma^2+\sigma^2}$ with an appropriate sum of their derivative orders. This does not hold when $\xi$ is non-linear, but our experience is that resulting scale of $\text{grad}^2$ is close to $\sqrt{\gamma^2+\sigma^2}$.

\section{Experiments}
In this section, we will compare LON \eqref{eq:lon} with CNN \eqref{eq:cnn}. To focus on the inner parts of the networks, i.e., we will set $W_j=\delta$, where $\delta$ is the dirac delta function.

To investigate the empirical difference between CNN and LON, we have considered the following cases: First we investigate their ability to estimate the gradient magnitude squared. Then we investigate their ability to estimate the area and perimeter length of random shapes and to classify these in groups. Finally, we perform a saliency analysis to investigate their ability to explain their results on the area and perimeter classification task.

\subsection{Estimating the gradient magnitude squared}
\begin{figure}
\centering
\subfigure[Original]{\label{fig:original}\includegraphics[width=0.3\linewidth]{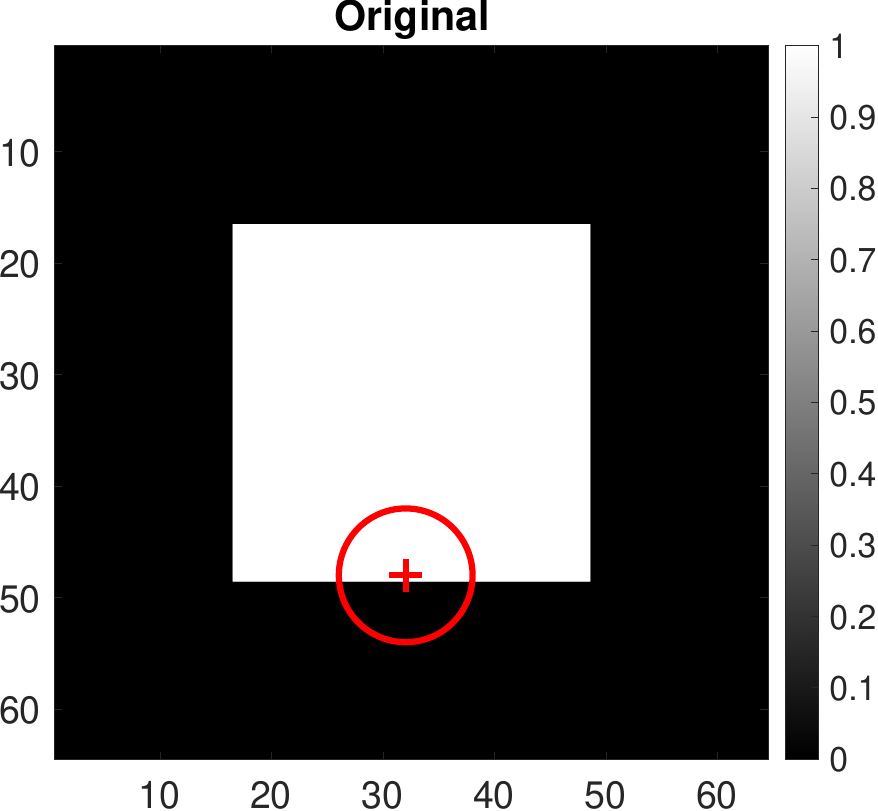}}
\hfill
\subfigure[Histogram of y-derivative]{\label{fig:hIr}\includegraphics[width=0.3\linewidth]{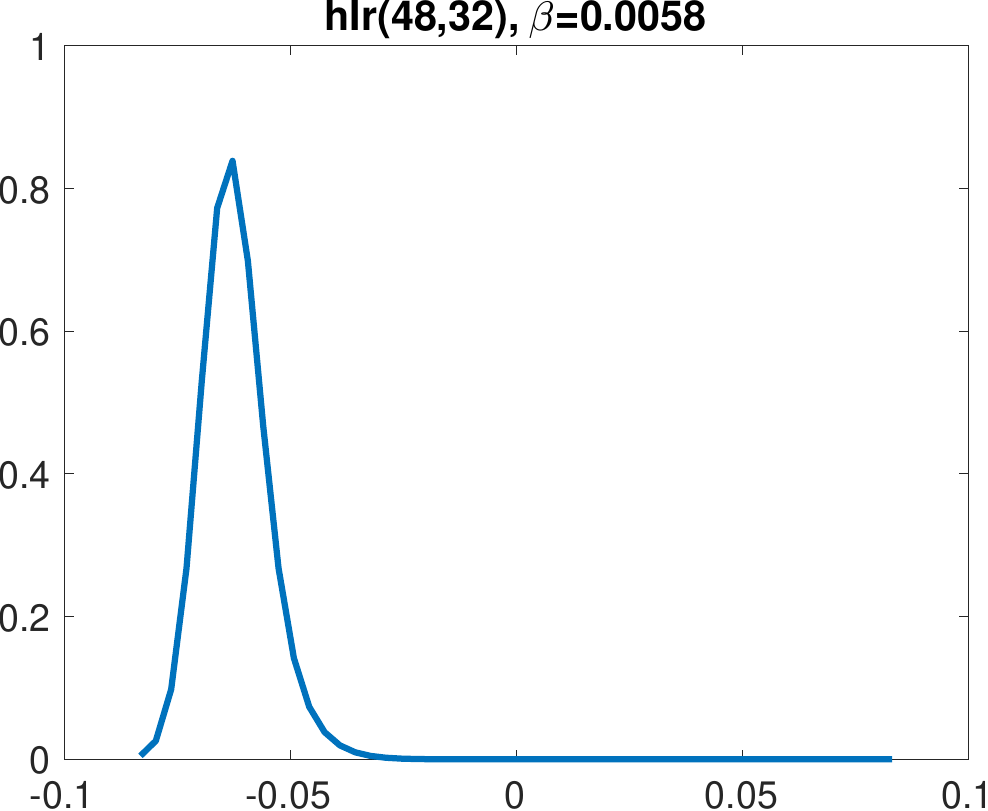}}
\hfill
\subfigure[Histogram of x-derivative]{\label{fig:hIc}\includegraphics[width=0.3\linewidth]{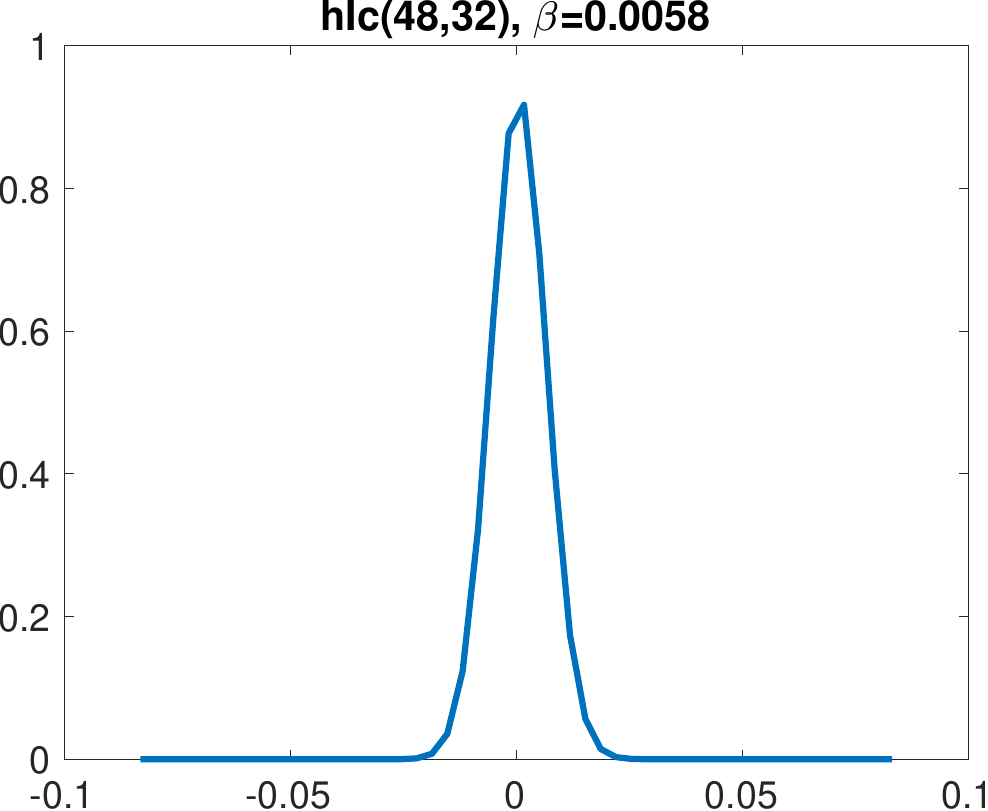}}
\subfigure[Gradient magnitude]{\label{fig:G}\includegraphics[width=0.3\linewidth]{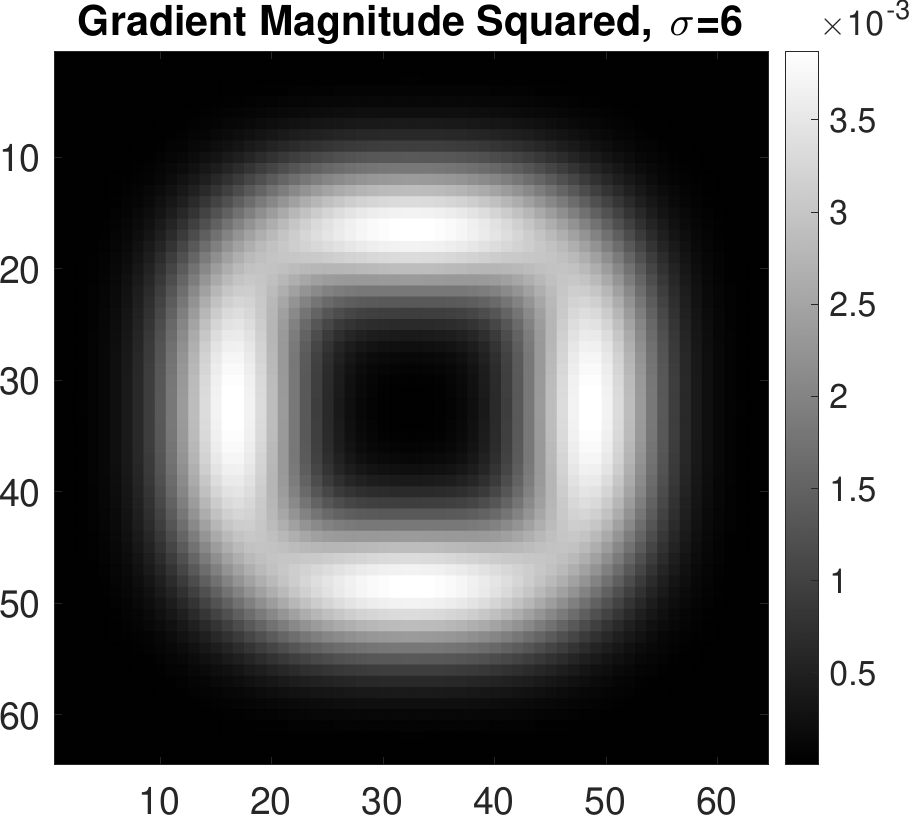}}
\hspace*{3mm}
\subfigure[LON]{\label{fig:Gh}\includegraphics[width=0.3\linewidth]{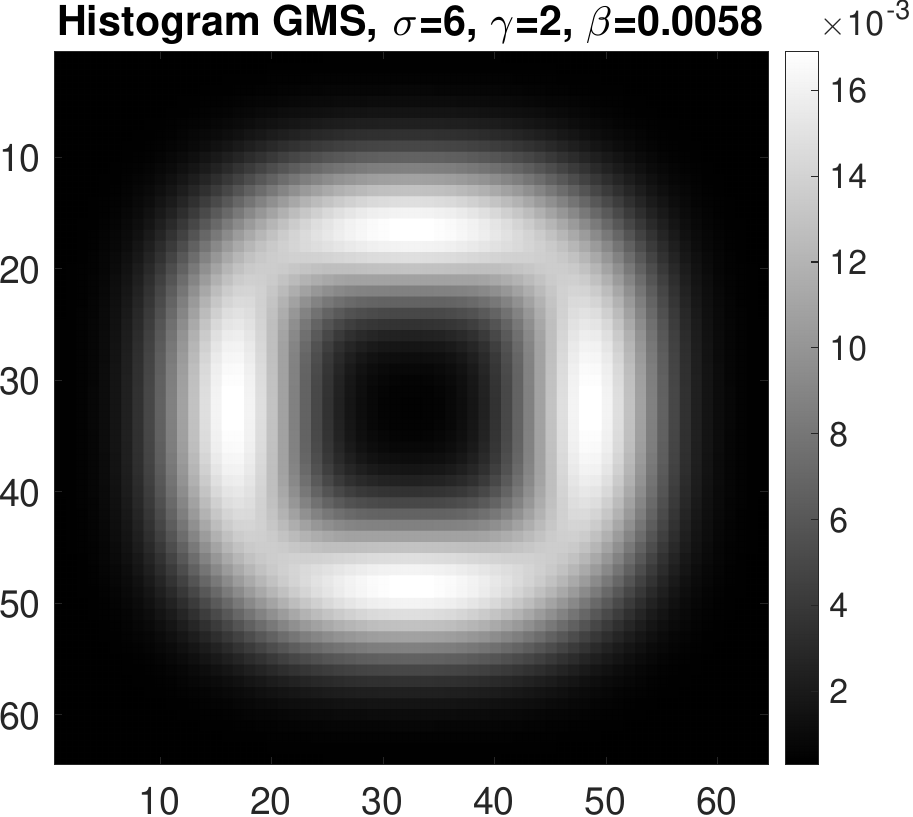}}
\caption{Comparing LON with the gradient magnitude%
. The original image (\subref{fig:original}) and its local histogram of the vertical and horizontal derivative (\subref{fig:hIr}) and (\subref{fig:hIc}) at the red mark and with a smoothing kernel of width indicated by the red circle.
(\subref{fig:G})
and (\subref{fig:Gh}), the gradient magnitude squared $\left|I(\vec{x})\right|^2$
and the locally orderless network \eqref{eq:grad3} with $A=\text{id}$.}
\label{fig:gradientMagnitude}
\end{figure}

The gradient magnitude squared has been used in image processing for decades as a rotational invariant indicator of the apparent edge of object-parts. Using spatial coordinates $\vec{x}=[x,y]$, then a direct implementation is given by
$\left|I(\vec{x})\right|^2 = \left(\frac{\partial I(\vec{x})}{\partial x}\right)^2+\left(\frac{\partial I(\vec{x})}{\partial y}\right)^2$.
Examples of $\left|I(\vec{x})\right|^2$ and \eqref{eq:grad3} on a simple image are shown in \Cref{fig:gradientMagnitude}.
We see that LON produces results which are visually very similar to the direct implementation of the gradient magnitude.

\begin{figure}
\centering
\subfigure[Input Images]{
\label{fig:mnistOriginal}
\begin{tikzpicture}
\node [
    above right,
    inner sep=0] (image) at (0,0) {\includegraphics[width=0.45\linewidth,trim=0mm 0mm 24mm 0mm,clip]{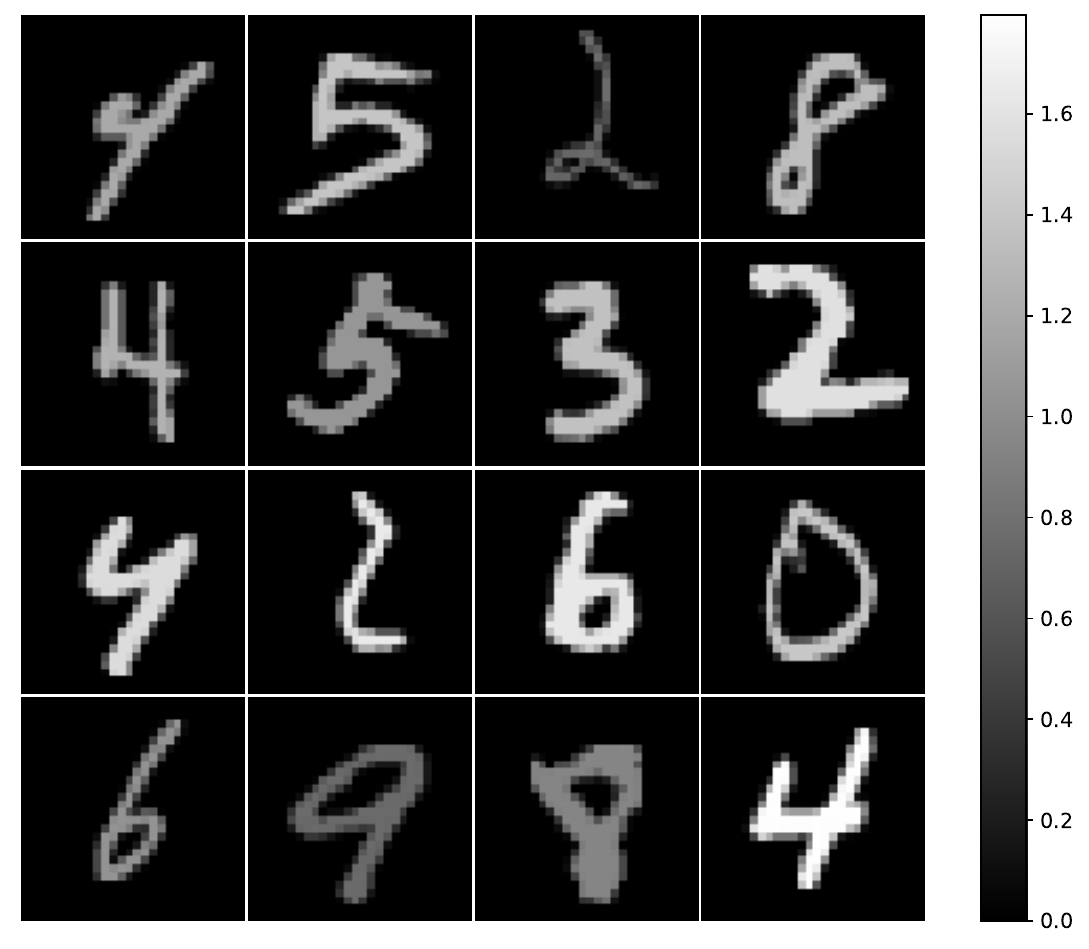}};
\end{tikzpicture}
}\hspace*{0mm}
\subfigure[Target]{
\label{fig:mnistTarget}
\begin{tikzpicture}
\node [
    above right,
    inner sep=0] (image) at (0,0) {\includegraphics[width=0.45\linewidth,trim=0mm 0mm 24mm 0mm,clip]{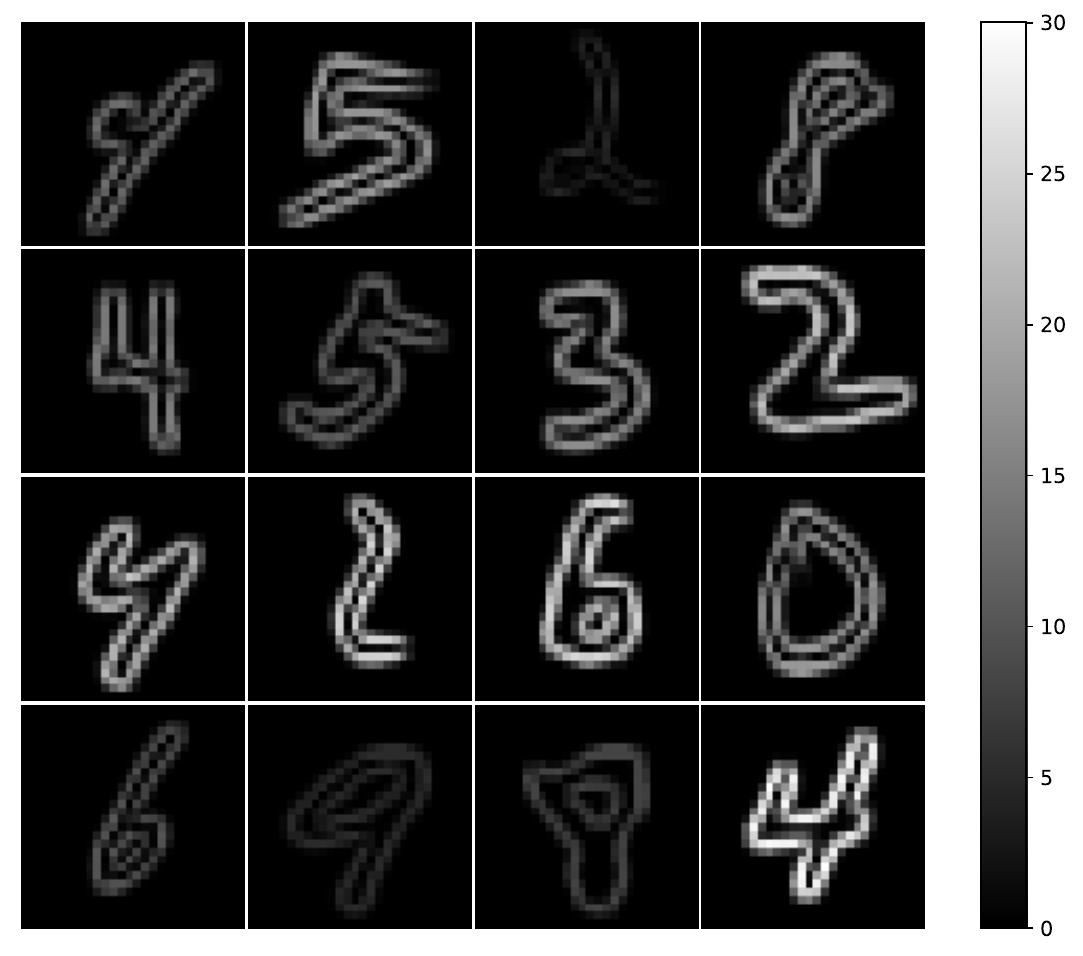}};
\begin{scope}[
    x={($0.1*(image.south east)$)},
    y={($0.1*(image.north west)$)}]
    \draw[latex-, very thick,red] (9,7) -- ++(1,1);
\end{scope}
\end{tikzpicture}
  }\\
\subfigure[LON, 2 kernels, 2 bins]{
\label{fig:mnistLON2bin}
\begin{tikzpicture}
\node [
    above right,
    inner sep=0] (image) at (0,0) {\includegraphics[width=0.45\linewidth,trim=0mm 0mm 24mm 0mm,clip]{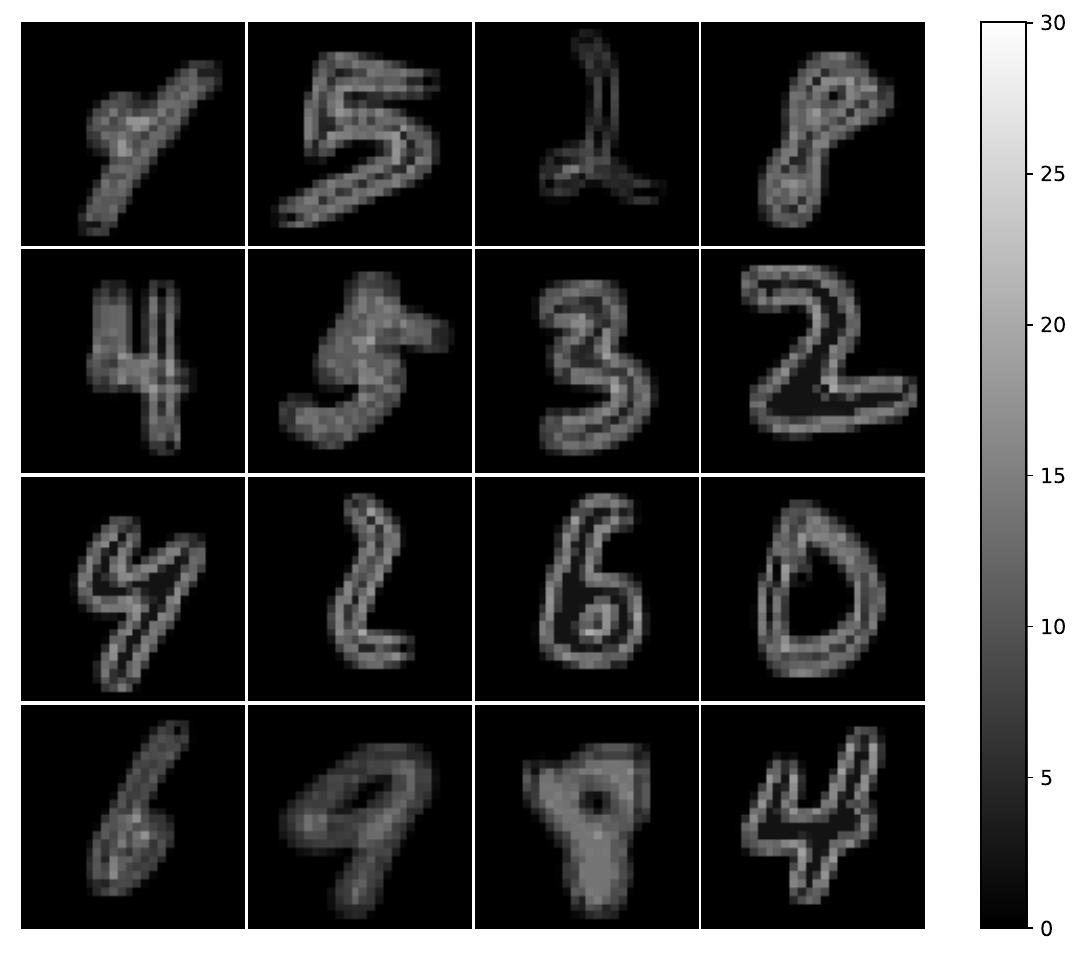}};
\begin{scope}[
    x={($0.1*(image.south east)$)},
    y={($0.1*(image.north west)$)}]
    \draw[latex-, very thick,red] (9,7) -- ++(1,1);
\end{scope}
\end{tikzpicture}
}\hspace*{0mm}
\subfigure[LON, 2 kernels, 8 bins]{
\label{fig:mnistLON8bin}
\begin{tikzpicture}
\node [
    above right,
    inner sep=0] (image) at (0,0) {\includegraphics[width=0.45\linewidth,trim=0mm 0mm 24mm 0mm,clip]{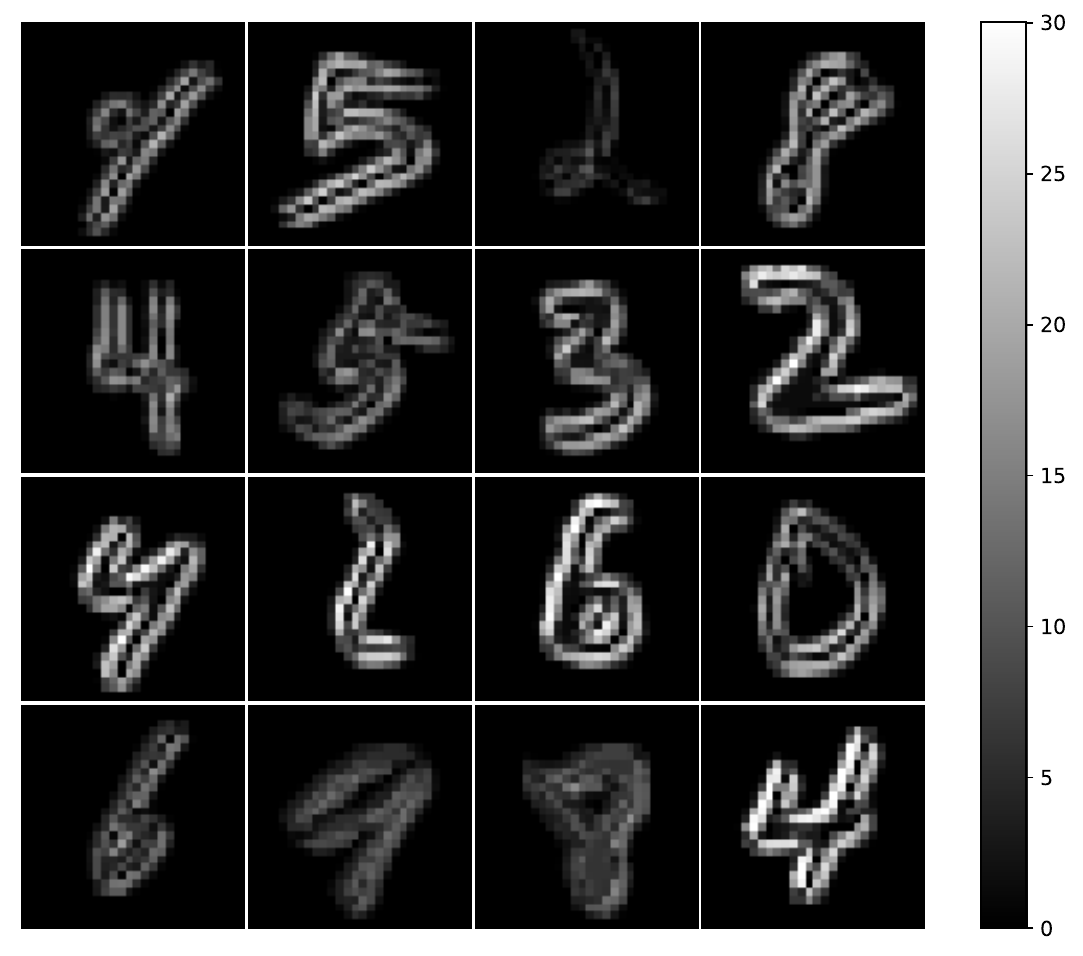}};
\begin{scope}[
    x={($0.1*(image.south east)$)},
    y={($0.1*(image.north west)$)}]
    \draw[latex-, very thick,red] (9,7) -- ++(1,1);
\end{scope}
\end{tikzpicture}
}\\
\subfigure[CNN, 2 kernels, ReLU]{
\label{fig:mnistCNNrelu}
\begin{tikzpicture}
\node [
    above right,
    inner sep=0] (image) at (0,0) {\includegraphics[width=0.45\linewidth,trim=0mm 0mm 24mm 0mm,clip]{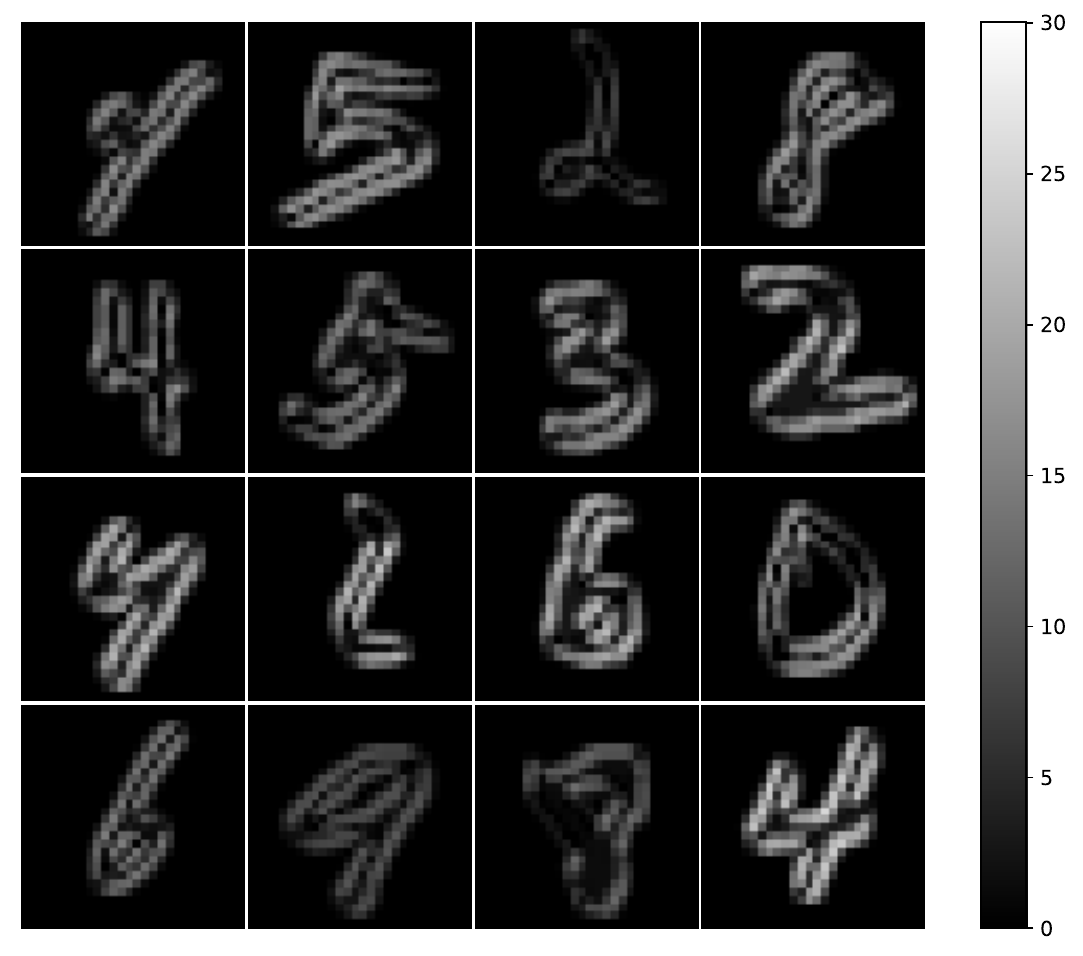}};
\begin{scope}[
    x={($0.1*(image.south east)$)},
    y={($0.1*(image.north west)$)}]
    \draw[latex-, very thick,red] (9,7) -- ++(1,1);
\end{scope}
\end{tikzpicture}
}\hspace*{0mm}
\subfigure[CNN, 2 kernels, sigmoid]{
\label{fig:mnistCNNsigmoid}
\begin{tikzpicture}
\node [
    above right,
    inner sep=0] (image) at (0,0) {\includegraphics[width=0.45\linewidth,trim=0mm 0mm 24mm 0mm,clip]{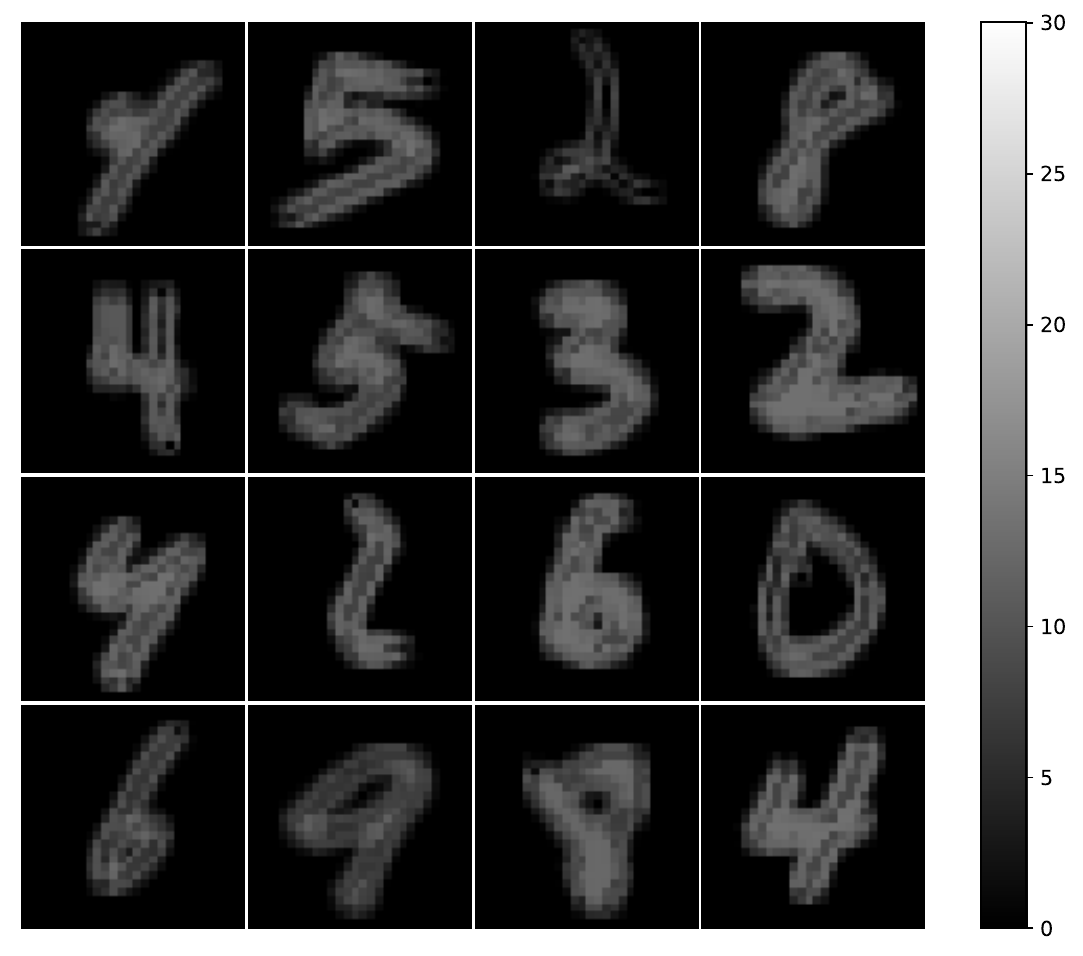}};
\begin{scope}[
    x={($0.1*(image.south east)$)},
    y={($0.1*(image.north west)$)}]
    \draw[latex-, very thick,red] (9,7) -- ++(1,1);
\end{scope}
\end{tikzpicture}
}
\caption{CNN achieves similar behavior as LON, but while LON succeeds in all directions, CNN does not.}
\label{fig:mnistgrad_demos}
\end{figure}

To compare the ability of CNN and LON to learn the gradient magnitude squared, we have calculated the gradient magnitude squared using $\left|I(\vec{x})\right|^2$ for objects from the MNIST database \cite{deng12}. Our hypothesis is that for a two-kernel system, both CNN and LON will learn a set of orthogonal derivative kernel, but in contrast to LON, CNN will not be able to learn the square nature of the gradient magnitude. To highlight the difference, the intensities of each hand-written characters is multiplied by a random scalar sampled from the continuous uniform distribution from $0.5$ to $2.0$, examples of which are shown in \Cref{fig:mnistOriginal} and \Cref{fig:mnistTarget}.
The networks all consists of a $3 \times 3$ convolution with their respective activation functions, followed by a $1 \times 1$ convolution. This results in networks with $21-35$ parameters in total. All networks are trained for $2000$ epochs with batch size $2048$, using Adam with learning rate $0.005$ and pixel-wise mean squared error. All models converged fast with similar learning curves.
Examples of results for both CNN and LON are shown in \Cref{fig:mnistgrad_demos} for a varying number of kernels, $f$, and the number of regular samples on the bias axis $i$. For both the CNN experiments, we see that edges at certain angles are not modeled correctly, while LON with 2 kernels successfully captures the edges in all directions, and LON with 8 kernels also captures the intensity variation accurately as reflected in the loss. The CNN with sigmoid activation appears to have significant difficulties with intensities and fails to predict the edges, particularly at certain orientations, (example is indicated by the red arrow
in figure \ref{fig:mnistgrad_demos})  

\subsection{Area versus perimeter regression and classification}
For our second set of experiments, we consider 2 simple problems: Regression and classification. We hypothesize that the structure of CNN will excel, when working with areas in $K*I$, due to its ReLU activation function, while LON will excel for isophotes (intensity level lines) in $K*I$, since it essentially operates on histogram bins.

\begin{figure}
\centering
\subfigure[]{\label{fig:smoothedNoise}\includegraphics[width=0.32\linewidth,trim=6.4cm 0.8cm 13.0cm 0cm,clip]{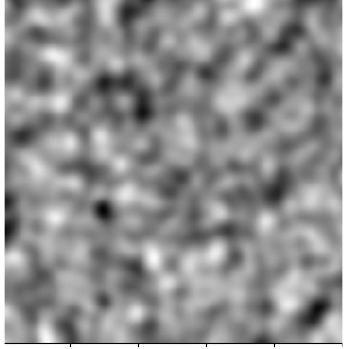}}
\subfigure[]{\label{fig:objectClipping}\includegraphics[width=0.32\linewidth,trim=19.4cm 0.8cm 0cm 0cm,clip]{ShapeGeneration.pdf}}
\subfigure[]{\label{fig:noisyClippings}\includegraphics[width=0.32\linewidth,trim=1.4cm 1cm 1cm 1cm,clip]{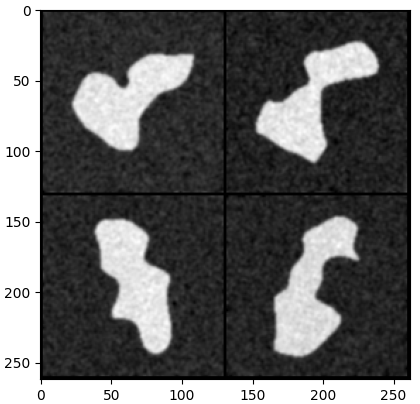}}
    \caption{Process for random shape generation: An image of iid.\ normal noise smoothed with a Gaussian kernel \subref{fig:smoothedNoise}, 
    its threshold and similar components \subref{fig:objectClipping}, and examplar objects with added iid.\ noise \subref{fig:noisyClippings}.}
    \label{fig:shape_generation}
\end{figure}

To test our hypothesis, we constructed a stochastic source of objects with varying area and perimeter, by generating random $512\times 512$ images from identically and independently distributed (iid.) normal noise. A Gaussian filtering was then applied with $\sigma=10$. The foreground areas are then selected as having intensity values $>75\% $ quantile. With the thresholded image, we then run connected component decomposition to separate each random shape while ignoring the incomplete shapes near boundaries as well as too small or too large shapes.  We finally place such shape into an image of fixed size $128\times 128$. The process in illustrated in \Cref{fig:shape_generation}.

\begin{figure*}
\centering
\setlength{\tabcolsep}{1mm}
\setlength{\tabcolsep}{1mm}
\begin{tabular}{m{0.4\linewidth}m{0.4\linewidth}}
     \hspace*{32mm}\footnotesize Perimeter Regressor & \hspace*{35mm}\footnotesize Area Regressor\\
\subfigure[]{\label{fig:AccPerimeterNoiseFree}\includegraphics[width=\linewidth,trim=0mm 2mm 13mm 14mm, clip]{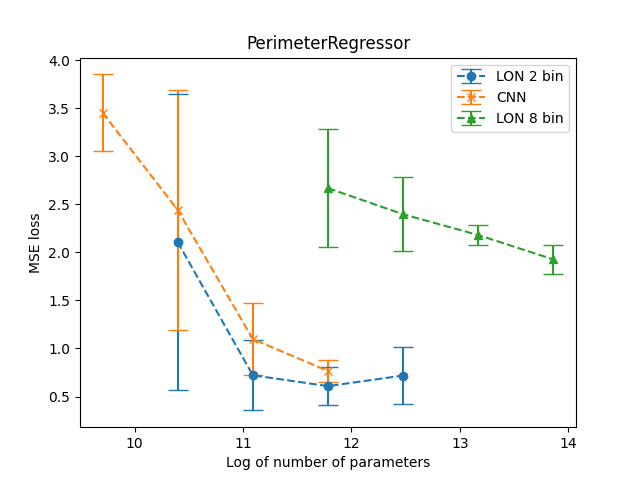}}
&
\subfigure[]{\label{fig:AccArea1NoiseFree}\includegraphics[width=\linewidth,trim=0mm 2mm 13mm 14mm, clip]{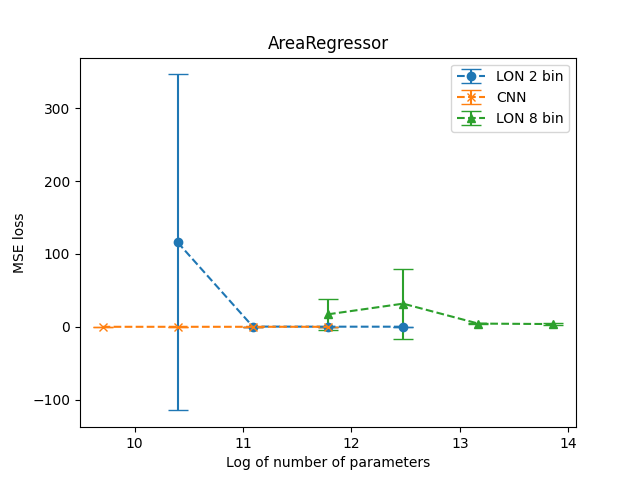}}
\end{tabular}
\caption{The mean square error by the logarithm of the number of parameters for CNN with 2 kernels and the ReLU, LON with 2 kernels and 2 bins and 8 bins on the regression task on the perimeter and area of random shapes without iid.\ noise (\Cref{fig:shape_generation}). Lower is better.}
\label{fig:regression}
\end{figure*}

For the regression task, we tested networks with varying number of kernels), and the results are shown in \Cref{fig:regression} on noiseless images of random objects.
Note that the number of parameters varies for CNN and LON by the number of kernels and bins. With an image with $|\Omega|$ pixels, $M$ kernels of $|K|$ pixels, CNN has $M|\Omega|+M|K|$ parameters. On the other hand, LON further has $N$ bins and thus $NM|\Omega|+M|K|$ parameters. In direct comparison, the number of parameters is dominated by the $|\Omega|$, and hence, LON is $N$ times larger than a CNN with the same number of kernels. However, if the CNN is given $NM$ kernels to compare with a LON with $N$ bins and $M$ kernels, then the LON has $(N-1)M|K|$ fewer parameters. In our experiments, we compared the network's performance on a logarithmic scale, where the subtle difference in the number of parameters is not highlighted. The models were trained using Adam optimizer, and we tuned the learning rate of both $1\cdot 10^{-3}$ and $5\cdot 10^{-4}$ for experiments and only the best results are reported.

For the regression task, we see wrt.\ the perimeter, the LON outperforms CNN in the 2-bin case in terms of the number of parameters used, however, for the 8-bin case it seems that the LON overfits. Wrt.\ area, LON with 2 bins has trouble converging, while CNN outperforms both LON. 

\begin{figure*}
\centering
\setlength{\tabcolsep}{1mm}
\begin{tabular}{m{0.01\linewidth}m{0.4\linewidth}m{0.4\linewidth}}
     & \hspace*{8mm}\footnotesize Perimeter Classifier & \hspace*{10mm}\footnotesize Area Classifier\\
     \hspace*{-1mm}\raisebox{0mm}{\rotatebox{90}{\footnotesize Large training set}} 
     & 
     \subfigure[]{\label{fig:AccPerimeterLarge}\includegraphics[width=\linewidth,trim=0mm 2mm 13mm 13mm, clip]{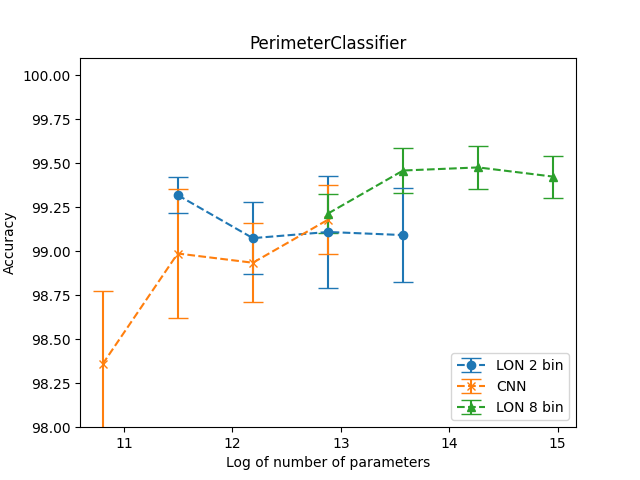}}
     &
     \subfigure[]{\label{fig:AccAreaLarge}\includegraphics[width=\linewidth,trim=0mm 2mm 13mm 13mm, clip]{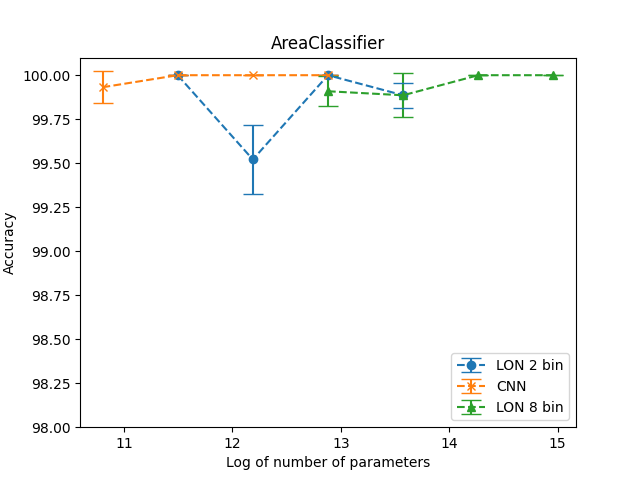}}\\
     \hspace*{-1mm}\raisebox{0mm}{\rotatebox{90}{\footnotesize Small training set}} 
     &
     \subfigure[]{\label{fig:AccPerimeterSmall}\includegraphics[width=\linewidth,trim=0mm 2mm 13mm 13mm, clip]{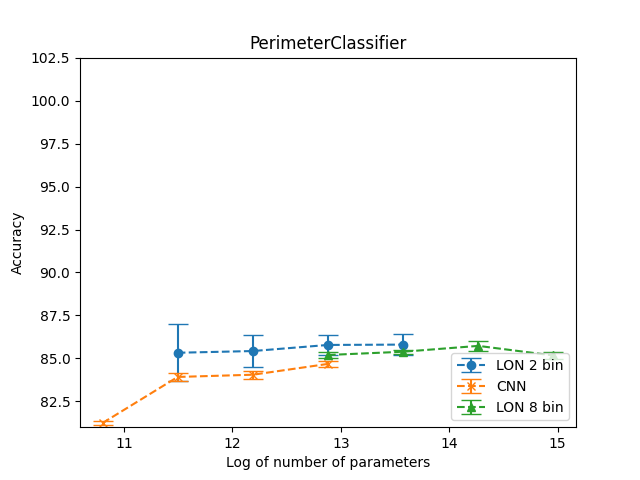}}
     &
     \subfigure[]{\label{fig:AccAreaSmall}\includegraphics[width=\linewidth,trim=0mm 2mm 13mm 13mm, clip]{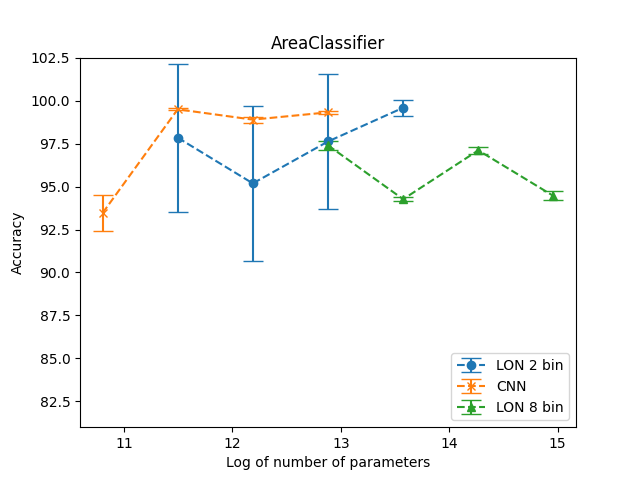}}
\end{tabular}
\caption{The accuracy by the logarithm of the number of parameters for CNN with 2 kernels and the ReLU, LON with 2 kernels and 2 bins and 8 bins on the the regression task on the perimeter and area of random shapes without noise but trained on many ($\approx$4000) or few ($\approx$1500) examples (\Cref{fig:shape_generation}). Higher is better.
}
\label{fig:classification}
\end{figure*}

For the classification task, we divided the objects into small, medium, and large wrt.\ either the area or perimeter length, and asked the network to correctly classify the three classes either by perimeter or area, while keeping the other constant. We also explored this problem in terms of a large or small training set. The results as a function of the number of parameters are shown in \Cref{fig:classification}.
Again it appears that LON with 2 bins is better at classifying objects in terms of their perimeter, while CNN is better at classifying objects wrt.\ area.
 
\subsection{Explainability by salience maps}
Explainability is an increasingly important property of machine learning algorithms, and as LON is linked to the boundary between apparent object parts, we hypothesize that saliency maps \cite{ittiModelSaliencybasedVisual1998} for LON will be more meaningful and thus easier interpreted than those of a CNN. We define saliency maps as
\begin{equation}
    \left|\frac{\partial E}{\partial I}\right|=\left|\frac{\partial E(Y,L(I))}{\partial I}\right|
\end{equation}
where $E$ is the error or loss function, $Y$ is the true class, $L$ is the network, and $I$ is the input image. The saliency maps express the gradient of each pixel wrt.\ to the similarity, thus what change in in the similarity a change in pixel value will cause. These are often considered the features of interest.

\begin{figure*}
\centering
    \begin{tabular}{|>{\centering}m{0.08\textwidth}|>{\centering}m{0.08\textwidth}|m{0.4\textwidth}|m{0.4\textwidth}|}
        \hline
        Type&Kernels&Perimeter classifier (constant area)&Area classifier (constant perimeter)\\\hline
        &2
        &{\includegraphics[width=0.39\textwidth,trim=1.9cm 4cm 1.5cm 3.5cm,clip]{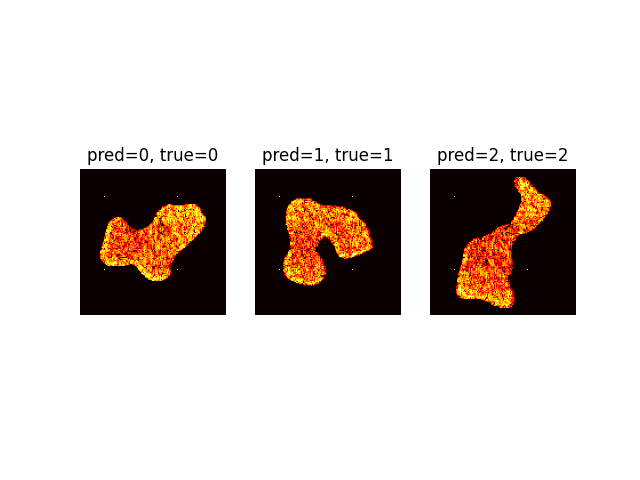}}
        &{\includegraphics[width=0.39\textwidth,trim=1.9cm 4cm 1.5cm 3.5cm,clip]{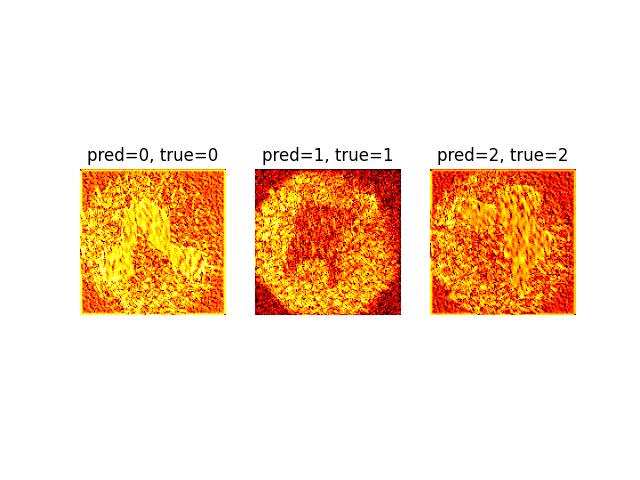}}\\
        \rotatebox{90}{CNN}&8
        &{\includegraphics[width=0.39\textwidth,trim=1.9cm 4cm 1.5cm 3.5cm,clip]{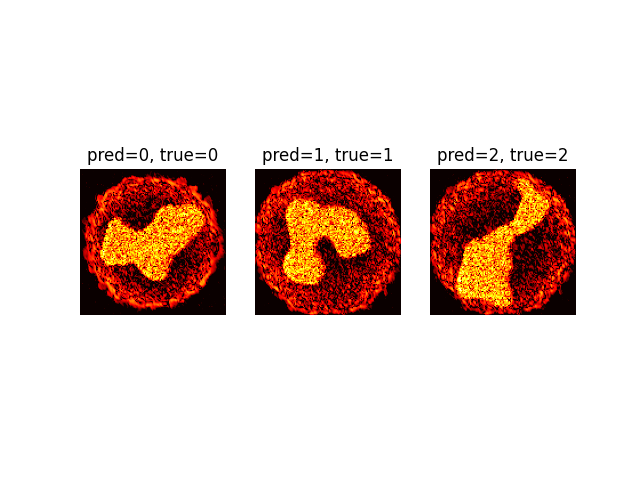}}
        &{\includegraphics[width=0.39\textwidth,trim=1.9cm 4cm 1.5cm 3.5cm,clip]{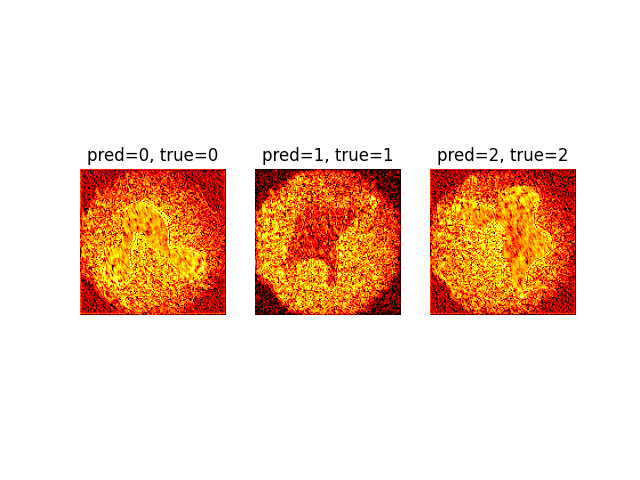}}\\\hline
        &2
        &{\includegraphics[width=0.39\textwidth,trim=1.9cm 4cm 1.5cm 3.5cm,clip]{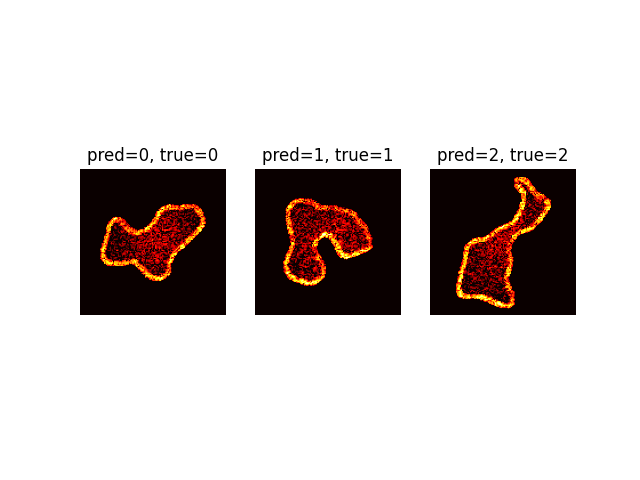}}
        &{\includegraphics[width=0.39\textwidth,trim=1.9cm 4cm 1.5cm 3.5cm,clip]{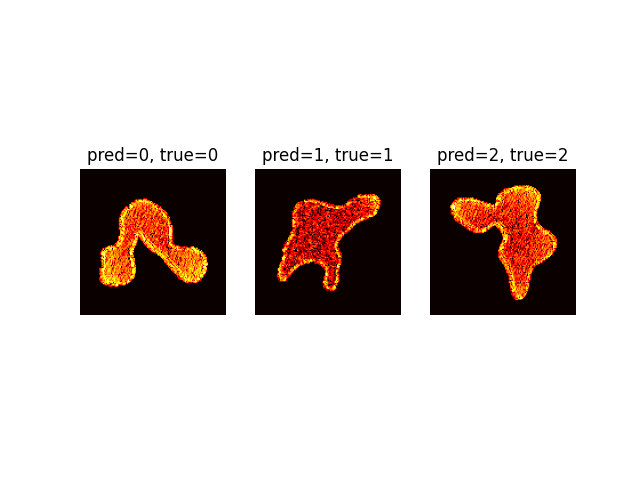}}\\
        \rotatebox{90}{LON 2 bins}&8
        &{\includegraphics[width=0.39\textwidth,trim=1.9cm 4cm 1.5cm 3.5cm,clip]{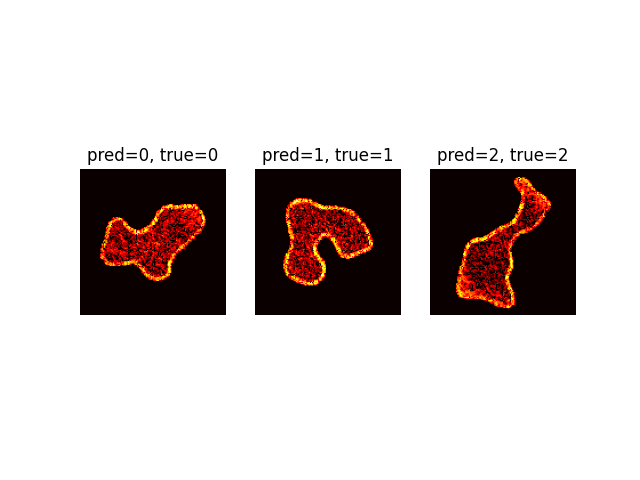}}
        &{\includegraphics[width=0.39\textwidth,trim=1.9cm 4cm 1.5cm 3.5cm,clip]{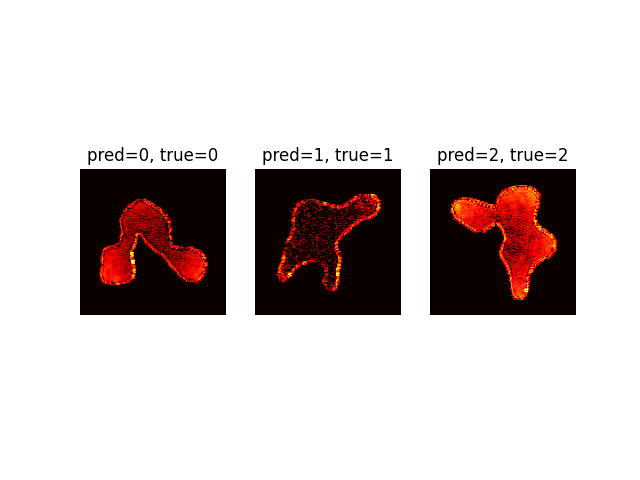}}\\\hline
        &2
        &{\includegraphics[width=0.39\textwidth,trim=1.9cm 4cm 1.5cm 3.5cm,clip]{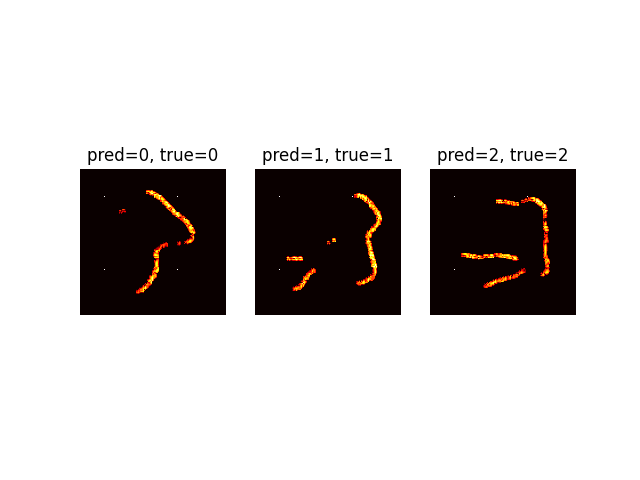}}
        &{\includegraphics[width=0.39\textwidth,trim=1.9cm 4cm 1.5cm 3.5cm,clip]{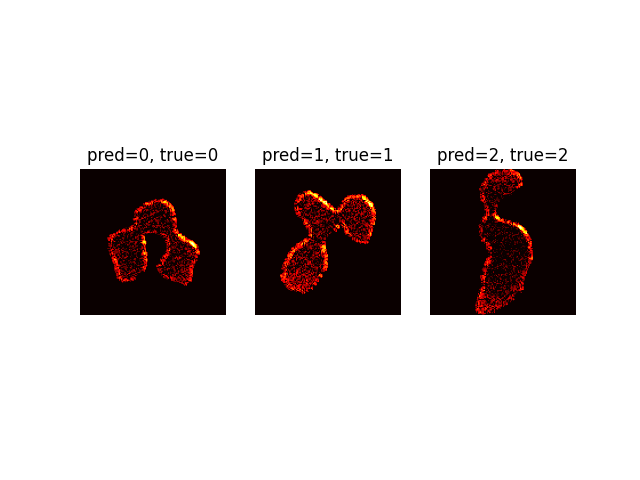}}\\
        \rotatebox{90}{LON 8 bins}&8
        &{\includegraphics[width=0.39\textwidth,trim=1.9cm 4cm 1.5cm 3.5cm,clip]{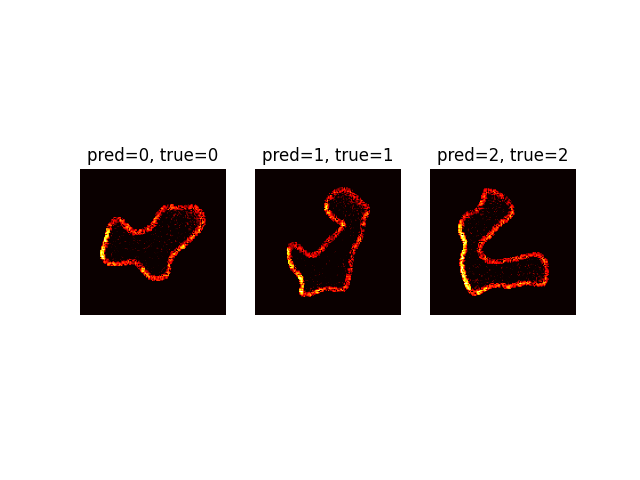}}
        &{\includegraphics[width=0.39\textwidth,trim=1.9cm 4cm 1.5cm 3.5cm,clip]{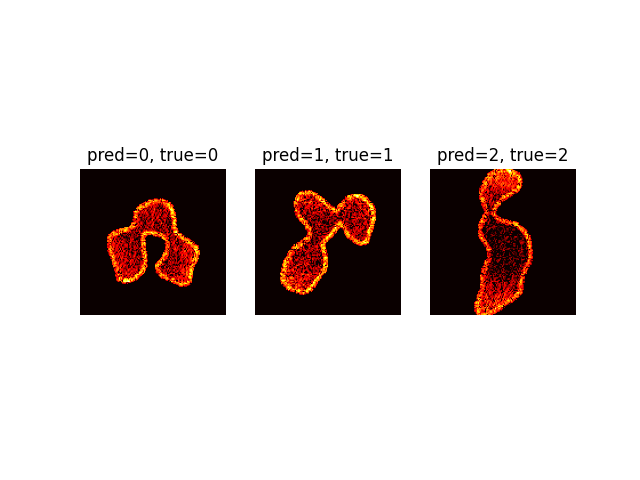}}\\\hline
    \end{tabular}
    \caption{Comparing saliency maps for perimeter and area classification on noise free images (\Cref{fig:shape_generation}). Shapes are grouped into 3 classes (pred=0..2), all networks uses 2 kernels, and light pixels have a large influence on the classification accuracy.}
    \label{fig:saliency}
\end{figure*}

In this experiment, we consider the classification task on random, noise free shapes, shown in \Cref{fig:noisyClippings}. The resulting saliency image is shown for various combinations of networks, channels, and the essential number of bins in \Cref{fig:saliency}. 
It shows that although all the models can make correct predictions with very high accuracy, as demonstrated in \Cref{fig:AccPerimeterLarge} and \Cref{fig:AccAreaLarge}, the saliency of the two models focusing on totally different regions. LON looks into the boundaries of the shape to get a perimeter estimator especially with 8 bins where CNN on the other hand is inferring the perimeter from both the foreground and the background.

For the case of area classification, the saliency map of CNN is still quite noisy and somewhat inconsistent in contrast to the LON which begins to make predictions based on the boundaries, but also takes the inside into account with more kernels or more bins. It seems that the saliency maps for LON are far more consistent across variation in parameters than for CNN and far better aligned with what humans would perceive as important for the two tasks.

\section{Discussion and Conclusion}
We have investigated the relation between local histograms and convolutional neural networks. We have investigated local histograms as defined in the locally orderless image framework, and we have shown that these histograms can be considered convolutional neural networks with a derivative of a sigmoid function as the activation function and when sampling the bias on a regular grid. We call this new type of network layer for locally orderless networks, and we have shown that with a simple added layer, convolutional and locally orderless networks can model each other. Due to the principle of the Unconscious statistician, this new type of network opens up new non-linear functions on the input, such as squaring, and thus, reveals new computation modes for convolutional type neural networks. We have demonstrated this for the gradient magnitude squared, which is an often used and rotational equivariant image operator, and we have shown that in contrast to a similar convolutional neural network, the locally orderless networks are both able to capture the rotational invariant nature of the gradient magnitude but also its amplitude. The sigmoid-type activation functions in convolutional neural networks are essentially thresholding functions, and their derivatives used in locally orderless networks are soft-indicator functions, and we, therefore, have hypothesized that the former will excel on tasks involving area operations, while the later on perimeter operations. Our experiments on both a regression and a classification task indicate that this is the case. Even more interestingly, we have examined the difference in saliency maps between the two network types for the area and perimeter classification experiments, and quite excitingly, the locally orderless images clearly depends on changes at the boundaries for both tasks while the convolutional neural network has a very diffuse saliency response. We see this as an important indication of locally orderless networks to be  better models for explainability as compared to convetional convolutional neural networks. For future work, we intend to investigate strategies for building large networks combining existing network technology with our new locally orderless networks, and we are particularly excited about investigating these new network types on segmentation tasks.

\bibliographystyle{splncs04}
\bibliography{refs.bib}

\end{document}